\documentclass{INTERSPEECH2023}

\usepackage{float}
\usepackage{cite} %
\interspeechcameraready

\title{Cross-Lingual Transfer Learning for Phrase Break Prediction with Multilingual Language Model}
\name{Hoyeon Lee, Hyun-Wook Yoon, Jong-Hwan Kim, Jae-Min Kim}
\address{
  NAVER Cloud, South Korea
}
\email{yeon.lee@navercorp.com}

\begin{document}
\newcommand{\hyunwookedit}[1]{\textcolor{blue}{#1}}
\newcommand{\yeonedit}[1]{\textcolor{green}{#1}}
\maketitle
 
\begin{abstract} %
Phrase break prediction is a crucial task for improving the prosody naturalness of a text-to-speech (TTS) system.
However, most proposed phrase break prediction models are monolingual, trained exclusively on a large amount of labeled data.
In this paper, we address this issue for low-resource languages with limited labeled data using cross-lingual transfer.
We investigate the effectiveness of zero-shot and few-shot cross-lingual transfer for phrase break prediction using a pre-trained multilingual language model.
We use manually collected datasets in four Indo-European languages: one high-resource language and three with limited resources.
Our findings demonstrate that cross-lingual transfer learning can be a particularly effective approach, especially in the few-shot setting, for improving performance in low-resource languages.
This suggests that cross-lingual transfer can be inexpensive and effective for developing TTS front-end in resource-poor languages.
\end{abstract}
\noindent\textbf{Index Terms}: cross-lingual transfer, multilingual, phrase break prediction, text-to-speech front-end, language model

\section{Introduction}
The text processing front-end, such as text normalization, grapheme-to-phoneme (G2P), and phrase break prediction, has become a core part of modern text-to-speech (TTS) systems.
Many studies have shown that these text processing modules successfully leverage the naturalness of synthetic speech in various approaches, including traditional statistical methods~\cite{sun2001intonational, schmid2004new, qian2010automatic} and deep learning-based methods~\cite{zhang2019neural, zhang2020unified, liu2020exploiting, futamata2021phrase, kunevsova2022detection, menshikova2019prosodic}.
Recently, with the great success of BERT~\cite{devlin2019bert} in various natural language processing (NLP) tasks, most of the proposed works have adopted mainstream pre-trained language models (PLMs)~\cite{devlin2019bert, sanh2019distilbert, pires2019multilingual, jiao2020tinybert}.
These works have reported stable and robust performance on several downstream tasks of the TTS front-end, such as text normalization~\cite{zhang2019neural}, G2P conversion~\cite{zhang2020unified}, and phrase break prediction~\cite{liu2020exploiting, futamata2021phrase, kunevsova2022detection, menshikova2019prosodic}.
While PLMs have enabled significant advances in a wide range of TTS front-end tasks, a sizeable set of labeled, task-specific data is essentially required.

However, a well-structured, large-scale database that thereby affects the quality of models carries a tremendous cost in linguistic annotation, as well as notorious tediousness, and subjectivity~\cite{fort2016collaborative, lauscher2020zero}.
Even building it by language or task is more challenging in a practical way.
Current research on the TTS front-end is actively conducted on manually constructed, large-scale monolingual datasets or a handful of resource-rich languages, such as English, to address the data scarcity issue.
One potential approach to address this challenge is to leverage relatively low-resources via \textit{cross-lingual transfer} from a high-resource language. This approach can be effective in reducing the cost and time of data annotation for low-resource languages, while improving their performance~\cite{pires2019multilingual, conneau2019cross, conneau2020unsupervised, xue2021mt5, wu2019beto, artetxe2019massively, pfeiffer2020mad, nooralahzadeh2020zero}.

In this paper, we empirically investigate the effectiveness of \textit{zero-shot} and \textit{few-shot cross-lingual transfer} for phrase break prediction in four Indo-European languages: English, French, Spanish, and Portuguese.
We explore this question using DistilmBERT~\cite{sanh2019distilbert, pires2019multilingual}, which is one of the mainstream pre-trained multilingual language models to utilize multilingual representation space.
Our manually annotated dataset consists of one high-resource language and three other relatively low-resource languages, which are approximately 12\%-16\% in size compared to the English dataset.
We consider two different transfer learning settings and compare the performance of 1) \textit{zero-shot, cross-lingual} -- fine-tuning on one source language and testing on different target languages, and 2) \textit{few-shot, cross-lingual} -- fine-tuning on one source language and a few target language instances, and testing on the target language.
Our contributions are summarized as follows:
\begin{itemize}
    \item This is the first study to investigate the effectiveness of using a cross-lingual transfer learning framework based on a pre-trained multilingual language model for phrase break prediction in low-resource languages.
    \item To verify its efficacy in transfer settings, we conducted cross-lingual transfer experiments in both zero-shot and few-shot settings. The experimental results showed that zero-shot, cross-lingual models are not enough to resolve the lack of large-scale labeled data in their current state but that the few-shot models' performance was comparable to or slightly better than that of the monolingual models.
    \item We discovered that DistilmBERT, a pre-trained multilingual language model, can quickly leverage multilingual representation space, adapting its knowledge from high-resource languages to low-resource ones.
 \end{itemize}

\begin{table*}[]
  \caption{Distribution of the dataset. The number of utterances and their ratio to English per language, the average number of subwords per utterance, the average subword length, and the average phrasing information ratio in an utterance. The standard deviation is shown in parentheses.}
  \label{tab:datasets}
  \centering
  \begin{tabular}{lrrrrrrr}
    \toprule
    \textbf{Language} & \textbf{Utterances} & \textbf{Ratio}   & \textbf{Subwords} & \textbf{Subword length} & \textbf{AP} & \textbf{IP} & \textbf{SB} \\
    \midrule
    English           & 69,169              & 1.00             & 17.89 (11.42)     & 3.47 (2.17) & 84.2 & 7.7  & 8.1 \\
    French            & 8,322               & 0.12             & 18.17 (4.99)      & 3.21 (2.20) & 73.6 & 17.6 & 8.7 \\
    Spanish           & 11,101              & 0.16             & 21.47 (11.66)     & 3.35 (2.19) & 86.4 & 6.4  & 7.2 \\
    Portuguese        & 10,000              & 0.15             & 17.46 (4.22)      & 2.94 (1.78) & 88.2 & 3.4  & 8.4 \\
    \bottomrule
  \end{tabular}
\end{table*}

\section{Related Work}

\subsection{Phrase Break Prediction}
A phrase break prediction refers to a pause or boundary between phrases, mainly for intonational or breath reasons, and plays an important role in comprehending the structure of sentences.
In TTS systems, phrase break prediction can be one of the most important front-end modules~\cite{klimkov2018phrase} along with text normalization~\cite{zhang2019neural} and G2P~\cite{zhang2020unified}. %
Although previous studies have indicated that the use of a deep learning-based model architecture~\cite{liu2020exploiting, futamata2021phrase, kunevsova2022detection, menshikova2019prosodic} in phrase break prediction results in better performance than traditional methods such as a hidden Markov model (HMM) based model~\cite{schmid2004new}, most models are trained on a large-scale monolingual dataset.
Futamata et al.~\cite{futamata2021phrase} employed contextual text representations obtained from the BERT model pre-trained on Japanese Wikipedia, along with various linguistic features such as part-of-speech (POS) and dependency tree (DEP), and showed significant improvement compared to the conventional methods.
However, these approaches for building phrase break prediction models typically demand a substantial amount of labeled data, which can be a significant challenge for low-resource languages.
\subsection{Cross-lingual Transfer}
Most supervised learning-based models heavily rely on a large amount of labeled data to achieve optimal performance, which can be challenging due to the data scarcity issue.
To resolve this issue, cross-lingual transfer from high-resource to low-resource languages can be employed~\cite{schuster2018cross, pfeiffer2020mad}.
A cross-lingual transfer is a strategy by which a model can solve diverse downstream tasks from a resource-poor target language despite receiving only a few examples or none at all in that language during the training process.
Typically, a model is pre-trained using a large corpus of monolingual data in different languages, including more than 100 diverse languages, and can also make use of multilingual corpora. Then, the model is fine-tuned with a specific target language~\cite{lample2019cross}. %
In recent studies, cross-lingual transfer in NLP comes with techniques that rely on continuous cross-lingual representation spaces for different languages~\cite{ruder2019survey} or multilingual transformer models that are pre-trained on large, multilingual corpora through the language modeling objectives~\cite{pires2019multilingual, conneau2019cross}.
Due to the advantages of using a shared representation, multilingual language model-based zero-shot~\cite{wu2019beto, artetxe2019massively} or few-shot~\cite{lauscher2020zero} cross-lingual transfer studies are making progress.

\section{Dataset Description} %

To assess the cross-lingual transfer learning framework for phrase break prediction, we collected utterances from various domains in four languages: English (EN), French (FR), Spanish (ES), and Portuguese (PT).
These languages share typologically similar features: they belong to the Indo-European language family~\cite{dryer2013world}, use the Latin alphabet~\cite{grefenstette2000estimation}, and have the same subject-verb-object (SVO)~\footnote{
    There are seven categories of word order found among the various human languages: subject-verb-object (SVO), subject-object-verb (SOV), object-verb-subject (OVS), object-subject-verb (OSV), verb-object-subject (VOS), and verb-subject-object (VSO)~\cite{dryer2013world}.
} word order~\cite{dryer2013world, comrie1987world}.
To prevent bias towards specific domains in our datasets, we collected utterances from diverse sources, such as news articles, blog posts, community websites, and the spoken language corpus.
We then manually curated utterances containing sentences that satisfy the following criteria: 1) complete sentence structure, and 2) a minimum of 4 words and a maximum of 25 words per sentence.

Each utterance consists of several phrasing information, which mainly comprises an intonation phrase (IP), accent phrase (AP), and end of the sentence (SB). IP refers phonetic pause inserted between phrases, and AP refers to a delimitation.
Following the annotation method of Futamata et al.~\cite{futamata2021phrase}, we constructed a phrase break dataset in which phrasing information was manually annotated by a linguistic expert. Each silence between words and word transitions in the recorded speech of collected utterances was labeled as a phrase break.

Since some text preprocessing techniques for a particular language may degrade performance in other languages~\cite{kudo2018sentencepiece, adebara2022towards}, we applied only minimal preprocessing.
For tokenization, we used a shared WordPiece~\cite{wu2016google} vocabulary instead of language-specific tokenization.
Most of the utterances in our dataset around 69,000 are in English, whereas the other three languages have much fewer scripts, roughly 12\%-16\% of the English dataset, primarily due to the higher cost of the annotation required.
Each utterance consisted of an average of 18.27 (\textit{SD} = 10.58) subwords, and the average length of a subword was 3.38 (\textit{SD} = 2.14).
In total, our dataset contained 98,592 utterances with 1.8 million subwords in four languages.
Table~\ref{tab:datasets} shows the distribution of our dataset by language.

\section{Zero-Shot Transfer} %
First, we focus on zero-shot, cross-lingual transfer for phrase break prediction in four languages: training on one language and testing on other unseen languages.

\subsection{Experimental Setup}

From our manually-curated dataset, we randomly select 6,000, 200, and 200 utterances as training, validation, and test sets for each language, respectively.
We use the shared multilingual WordPiece vocabulary for utterances across all languages and take sequences of subwords as model input.

We follow the same architecture as DistilmBERT~\cite{sanh2019distilbert}, which is a distilled version of multilingual BERT (mBERT)~\cite{pires2019multilingual} trained on monolingual corpora in 104 languages, including English, French, Spanish, and Portuguese. In detail, the model consisted of 6 transformer layers with 12 heads in each layer, 768 hidden dimensions, making up 134M parameters\footnote{
    Note that DistilmBERT has relatively smaller parameters compared to the 177M parameters of mBERT-base and is about twice as fast as mBERT-base~\cite{sanh2019distilbert}. In our preliminary experiments, we confirmed that the distilled model was enough to effectively train for phrase break prediction, our downstream task, in various languages.
}.
DistilmBERT is fine-tuned with a maximum sequence length of 128 subword tokens to accommodate relatively long utterances for 5 epochs\footnote{
    DistilmBERT was trained on a single NVIDIA Tesla V100 GPU for every cross-lingual transfer experiment including subsequent few-shot settings.
}.
To select the best hyperparameters, we use the validation set performance.
We grid-search for the following hyperparameter values and select those in bold as our best: a batch size from \{\textbf{16}, 32\}, a learning rate of Adam optimizer \cite{kingma2014adam} from \{2e-5, \textbf{5e-5}, 5e-6\}, and a dropout rate of the last layer from \{\textbf{0.1}, 0.2\}.
The loss function is cross-entropy loss, and we measure the model performance using a macro-averaged F1 score.

\subsection{Results and Analysis}
Table~\ref{tab:zero-shot} presents the macro-averaged F1 scores for different training and test languages, with rows and columns representing each language, respectively. The numbers underlined  refer to the results in the monolingual setting.
As expected, the performance drops in zero-shot results for all target languages compared to the monolingual results, with the degree of the decline varying significantly across languages.
For example, when training on Spanish and testing on French, we observe the largest performance drop of 26.19\%. On the other hand, when training on Portuguese and testing on English, it is only 3.34\%, which is the smallest drop, and very close to the result of monolingual models.
Notably, given the monolingual results (underlined numbers in Table~\ref{tab:zero-shot}), training and testing in English produces the poorest performance compared to other languages, even though English has the highest proportion of pre-training corpora used in multilingual model training.
These results imply that zero-shot cross-lingual transfer, even  without using a single target language instance, can mitigate the data scarcity issue to some extent.
However, it is also clear that it cannot achieve the same level of performance as monolingual models, which could be a feasible alternative for real-world TTS front-end applications.
Therefore to further explore beyond the limitation of a zero-shot setting, we proceed to investigate cross-lingual transfer in few-shot settings in the following section.

\begin{table}[]
  \caption{Macro-F1 results of zero-shot cross-lingual transfer experiments. Monolingual results are \underline{underlined}, and the best results are \textbf{bolded}.}
  \label{tab:zero-shot}
  \centering
  \begin{tabular}{lcccc}
    \toprule
    Train \textbackslash Test   & EN                & FR                & ES                & PT                \\
    \midrule
    EN                          & \textbf{\underline{82.12}} & 76.98             & 72.16             & 86.39             \\
    FR                          & 73.71             & \textbf{\underline{92.37}} & 67.37             & 76.84             \\
    ES                          & 71.63             & 66.18             & \textbf{\underline{91.40}} & 70.38             \\
    PT                          & 78.78             & 76.72             & 68.37             & \textbf{\underline{92.62}} \\
    \bottomrule
  \end{tabular}
\end{table}
 
\section{Few-Shot Transfer}
To improve the inadequate results in zero-shot transfer, we now investigate the effectiveness of few-shot, cross-lingual transfer with a modest number of target language instances in conjunction with the source language data. Our objective is to narrow the performance gap with the monolingual model.

\subsection{Experimental Setup}
Unlike in zero-shot transfer, we employ a two-step fine-tuning approach in this setting.
For the first fine-tuning step, we consider two different scenarios -- \textit{English-Only} and \textit{Augmented} -- to analyze each few-shot, cross-lingual transfer performance.
In both scenarios, we expand a source language dataset in English to 60,000 examples to take advantage of the abundance of available data.
For the first scenario (\textit{English-Only}), we use one of the general cross-lingual transfer methods, which fine-tunes the model solely on a large amount of English data.
Additionally, we consider another scenario (\textit{Augmented}) based on the generally accepted correlation between larger dataset size and better performance~\cite{delobelle2020robbert}.
In this scenario, English and all other languages except for the target language are included during the training.
Note that the datasets used for training, validation, and testing are identical to those used in the zero-shot experiment.
This approach reflects a realistic scenario that utilizes all available datasets.

After the first fine-tuning step, the second fine-tuning step is conducted in the same manner for both scenarios.
During this step, we progressively increase the size of the target language examples \textit{k} in powers of 2 as follows: $k \in {\{4, 8, 16,..., 4096\}}$.
Note that the model architecture and hyperparameters used in this experiment are the same as those in the zero-shot setting.

\subsection{Results and Analysis}

\begin{figure}[]
  \centering
  \includegraphics[width=\linewidth]{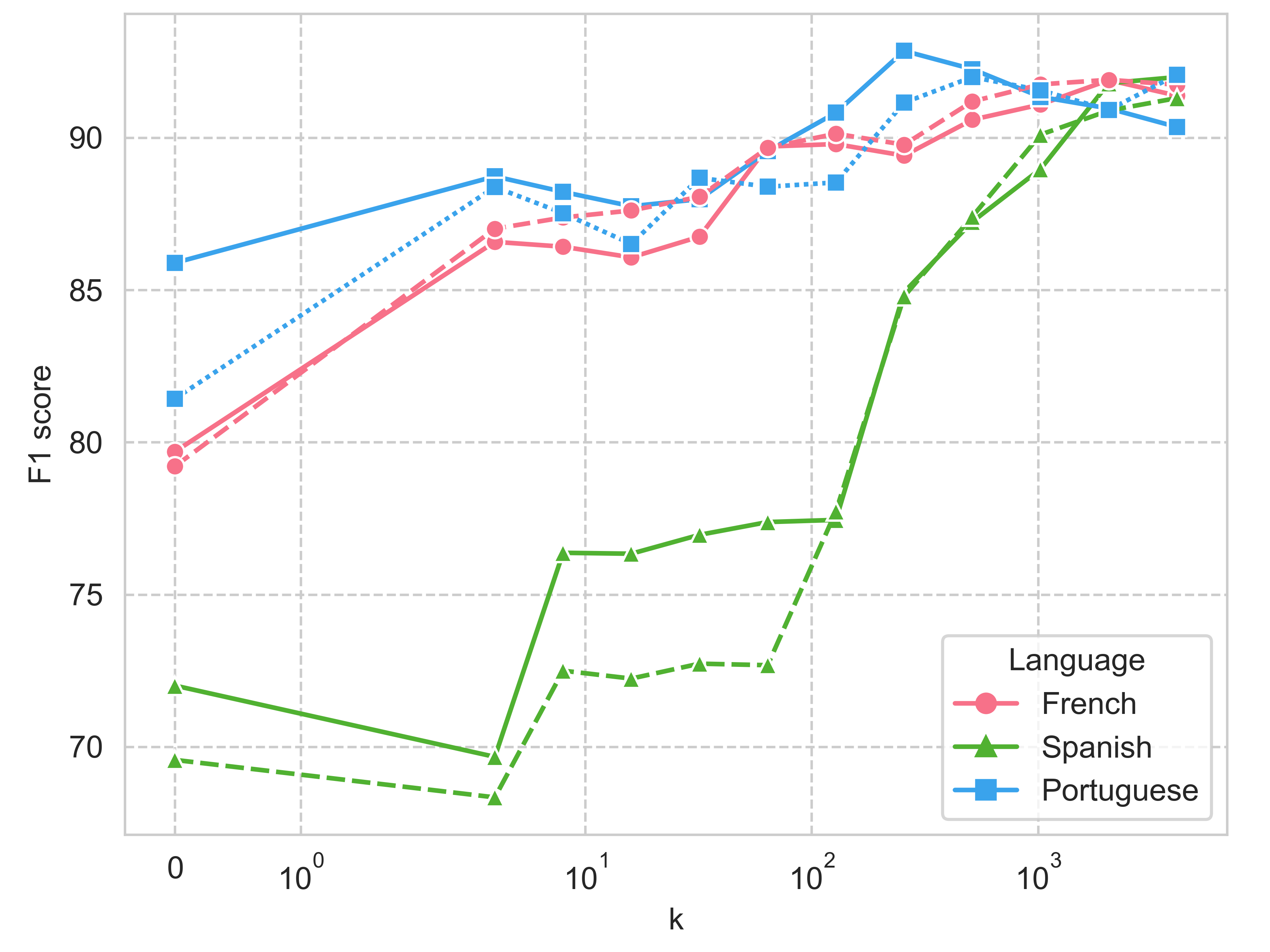}
  \caption{Results of few-shot, cross-lingual transfer experiments with varying numbers of target language examples k. The solid lines indicate the English-Only scenario, and the dashed lines indicate the Augmented scenario.}
  \label{fig:few-shot-results}
\end{figure}

\begin{table*}[]
    \caption{Results of few-shot, cross-lingual transfer in the English-Only scenario. k denotes the number of target language instances, score denotes a macro-averaged F1 score, and ${\Delta}$ denotes difference with respect to the zero-shot setting.}
    \label{tab:few-shot}
    \centering
    \begin{tabular}{lcrrrrrrrrrrrr}
        \toprule
        & \textit{zero-shot} & \multicolumn{2}{c}{\textit{k=8}} & \multicolumn{2}{c}{\textit{k=32}} & \multicolumn{2}{c}{\textit{k=128}} & \multicolumn{2}{c}{\textit{k=256}} & \multicolumn{2}{c}{\textit{k=512}} & \multicolumn{2}{c}{\textit{k=2048}} \\
        Language & score    & score & ${\Delta}$ & score & ${\Delta}$ & score & ${\Delta}$ & score & ${\Delta}$ & score & ${\Delta}$ & score & ${\Delta}$ \\
        \midrule
        FR       & 79.69    & 86.42 & 6.73 & 86.75 & 7.06 & 89.79 & 10.1 & 89.41 & 9.72 & 90.59 & 10.9 & \textbf{91.89} & 12.2 \\
        ES       & 72.01    & 76.37 & 4.36 & 76.96 & 4.95 & 77.45 & 5.44 & 84.93 & 12.92 & 87.24 & 15.23 & \textbf{91.79} & 19.78 \\
        PT       & 85.89    & 88.22 & 2.33 & 87.98 & 2.09 & 90.82 & 4.93 & \textbf{92.85} & 6.96 & 92.25 & 6.36 & 90.96 & 5.07 \\
        \bottomrule
    \end{tabular}
\end{table*}

\begin{table*}[]
    \caption{Results of few-shot, cross-lingual transfer in the Augmented scenario. k denotes the number of target language instances, score denotes a macro-averaged F1 score, and ${\Delta}$ denotes difference with respect to the zero-shot setting.}
    \label{tab:few-shot-aug}
    \centering
    \begin{tabular}{lcrrrrrrrrrrrr}
        \toprule
        & \textit{zero-shot} & \multicolumn{2}{c}{\textit{k=8}} & \multicolumn{2}{c}{\textit{k=32}} & \multicolumn{2}{c}{\textit{k=128}} & \multicolumn{2}{c}{\textit{k=256}} & \multicolumn{2}{c}{\textit{k=512}} & \multicolumn{2}{c}{\textit{k=2048}} \\
        Language & score    & score & ${\Delta}$ & score & ${\Delta}$ & score & ${\Delta}$ & score & ${\Delta}$ & score & ${\Delta}$ & score & ${\Delta}$ \\
        \midrule
        FR       & 79.21    & 87.38 & 8.4 & 88.06 & 8.85 & 90.13 & 10.92 & 89.77 & 10.56 & 91.19 & 11.98 & \textbf{91.9} & 12.69 \\
        ES       & 69.57    & 72.5 & 2.93 & 72.73 & 3.16 & 77.73 & 8.16 & 84.79 & 15.22 & 87.41 & 17.84 & \textbf{90.89} & 21.32 \\
        PT       & 81.43    & 87.52 & 6.09 & 88.68 & 7.25 & 88.53 & 7.1 & 91.15 & 9.72 & \textbf{91.99} & 10.56 & 90.91 & 9.48 \\
        \bottomrule
    \end{tabular}
\end{table*}

The overall results of few-shot, cross-lingual transfer in both scenarios across all languages are presented in Figure~\ref{fig:few-shot-results}.
The results depicted in Figure~\ref{fig:few-shot-results} indicate consistent and significant improvements in performance across all languages and scenarios, as the number of target language instances represented by \textit{k} increases.
Even with very few instances, less than 100 in total, substantial performance gains are observed in comparison to zero-shot setting in all cases.
We can also observe that when trained on only a few target examples (i.e., from 4 to 256), the score rapidly increased during the initial stages, then gradually slowed down as more examples were added, and eventually converged.
Interestingly, in some cases, the performance even declined with a larger number of examples, particularly when \textit{k} was greater than 2,000.
Considering the results in Figure~\ref{fig:few-shot-results}, we conjecture that DistilmBERT, a pre-trained multilingual language model, can leverage multilingual representation space effectively and transfer its knowledge of the phrase break prediction task to a new unseen language.

\subsubsection{Results in English-Only Scenario}
Table~\ref{tab:few-shot} shows a detailed breakdown of the few-shot transfer results for the \textit{English-Only} scenario.
To highlight the performance changes with respect to \textit{k}, we included the zero-shot performance from a model exclusively fine-tuned on an English dataset, as well as the few-shot performance, and the differences between them.
As shown in Table~\ref{tab:few-shot}, the zero-shot transfer performance is enhanced in French and Portuguese and remains nearly the same in Spanish since the size of the English training set increases in the first fine-tuning step.
This implies that training on a larger dataset can lead to learning more knowledge about the phrase break prediction task.
We notice a consistent pattern in which performance gradually improves and achieves comparable results to a monolingual model trained with supervised learning.
Despite a marginal difference of 0.48, the French few-shot, cross-lingual model (91.89, \textit{k}=2,048) scored slightly lower than the monolingual model (92.37), whereas the Spanish and Portuguese models outperformed when \textit{k}=2,048 and \textit{k}=256, respectively.
Surprisingly, in the case of Portuguese, the model obtained the best F1 score when only 256 training instances were used.
Furthermore, the performance in Spanish improved to 91.99 even when \textit{k} was set to 4,096.

\subsubsection{Results in Augmented Scenario}
Table~\ref{tab:few-shot-aug} presents the results of few-shot transfer for the \textit{Augmented} scenario.
As in the previously presented findings, we can see a steady improvement as \textit{k} increases, following a similar pattern to the \textit{English-Only} scenario.
On the other hand, zero-shot performance generally decreased by less than 4\%. 
Interestingly, even with a relatively small amount of high-resource language data (as shown in Table~\ref{tab:zero-shot}), the performance in Spanish and Portuguese was lower than that of zero-shot transfer.
This demonstrates that the model is not able to generalize well across different languages during the initial fine-tuning step, as the datasets of different languages and sizes are trained simultaneously.
Consequently, this affects the next fine-tuning step, where the target language examples are added to the training set.
As you can see in Figure~\ref{fig:few-shot-results}, languages with the lower zero-shot performance show poorer performance than the \textit{English-Only} scenario.
A notable observation is that, as \textit{k} reaches a large size (greater than 1,000), comparable performance is reached, and even beyond that point, performance gradually improves.

\subsubsection{Analysis}
As we have conducted few-shot transfer experiments in two scenarios, we can verify that cross-lingual transfer is possible in a phrase break prediction task and report considerable gains with only a small amount of labeled data.
Since PLMs were introduced, training a large amount of labeled data has become more essential, as it always leads to better performance on various downstream tasks, including TTS front-end modules.
However, our findings demonstrate that few-shot transfer can fully address the lack of labeled data and narrow the gap with that, while zero-shot transfer, the most cost-effective approach that does not use any target example, cannot be used as it is.
Furthermore, the fact that a distilled version of the multilingual language model was enough to leverage its representation ability will be more beneficial for real-time TTS applications.

\section{Conclusion}
In this study, we present the first comprehensive research to verify the effectiveness of cross-lingual transfer for phrase break prediction in less-resourced languages using a pre-trained multilingual language model.
We investigate zero-shot and few-shot transfer learning settings to find an inexpensive alternative for low-resource language adaptation.
While our experimental results in the zero-shot setting did not rise to the performance of the monolingual model trained on a large-scale dataset with consistent labeling, they demonstrated the capability of the multilingual model to perform the transfers required for this task.
Moreover, we found that few-shot transfer using only a small number of annotated examples led to performance comparable to that of the monolingual model.
Our findings indicate that a few-shot, cross-lingual model can be an effective solution to address the challenge of limited annotated data in TTS front-end applications.
In future work, we plan to investigate the efficacy of cross-lingual transfer for other TTS front-end tasks, such as grapheme-to-phoneme (G2P) conversion, in languages with limited resources.
Furthermore, in future studies, we will explore whether the transfer learning approach works best even between typologically distinct languages, such as English and Korean.

\section{Acknowledgements}
This work was supported by Voice\&Avatar, NAVER Cloud, Seongnam, Korea.

\bibliographystyle{IEEEtran}
\bibliography{mybib}

\end{document}